%% file: egpaper_for_review.tex
\documentclass[10pt,twocolumn,letterpaper]{article}

\usepackage{cvpr}
\usepackage{times}
\usepackage{epsfig}
\usepackage{graphicx}
\usepackage{amsmath}
\usepackage{amssymb}
\usepackage{comment}
\usepackage{subcaption}
\usepackage{caption}
\usepackage{booktabs}
\usepackage{tabularx}
\usepackage{multirow}
\usepackage{mathtools}
\DeclarePairedDelimiter{\floor}{\lfloor}{\rfloor}
\usepackage{enumitem}
\setlist[itemize]{itemsep=0.05em, topsep=2pt}

\usepackage[pagebackref=true,breaklinks=true,letterpaper=true,colorlinks,bookmarks=false]{hyperref}

 \cvprfinalcopy 

\ifcvprfinal\fi
\begin{document}

\title{The Translucent Patch: A Physical and Universal Attack on Object Detectors}

\author{Alon Zolfi, Moshe Kravchik, Yuval Elovici, Asaf Shabtai\\
Department of Software and Information Systems Engineering\\
Ben-Gurion University of the Negev\\
{\tt\small \{zolfi,moshekr\}@post.bgu.ac.il, \{elovici,shabtaia\}@bgu.ac.il}
}

\maketitle

\input{sections/abstract}

\input{sections/introduction}

\input{sections/related_work}

\input{sections/method}

\input{sections/tesla}

\input{sections/evaluation}

\input{sections/conclusion}

{\small
\bibliographystyle{ieee_fullname}
\bibliography{egbib}
}

\end{document}

%% file: sections/abstract.tex
\begin{abstract}
Physical adversarial attacks against object detectors have seen increasing success in recent years.
However, these attacks require direct access to the object of interest in order to apply a physical patch.
Furthermore, to hide multiple objects, an adversarial patch must be applied to each object.
In this paper, we propose a contactless translucent physical patch containing a carefully constructed pattern, which is placed on the camera's lens, to fool state-of-the-art object detectors.
The primary goal of our patch is to hide all instances of a selected target class.
In addition, the optimization method used to construct the patch aims to ensure that the detection of other (untargeted) classes remains unharmed.
Therefore, in our experiments, which are conducted on state-of-the-art object detection models used in autonomous driving, we study the effect of the patch on the detection of both the selected target class and the other classes.
We show that our patch was able to prevent the detection of 42.27\% of all stop sign instances while maintaining high (nearly 80\%) detection of the other classes.
\end{abstract}

%% file: sections/introduction.tex
\section{\label{sec:intro}Introduction}

In recent years, deep neural networks (DNNs) and particularly convolutional neural networks (CNNs) have become a state-of-the-art solution for computer vision tasks, such as image classification~\cite{krizhevsky2012imagenet,he2015delving}, object detection~\cite{redmon2018yolov3,ren2015faster}, and image segmentation~\cite{chen2014semantic,chen2019hybrid}.
This is mainly due to DNNs' ability to accurately model complex multivariate data.
However, such models' effectiveness depends heavily on their robustness to attacks that target the model itself; i.e., adversarial learning attacks.

\begin{figure}[ht!]
    \centering
    \includegraphics[width=0.9\linewidth]{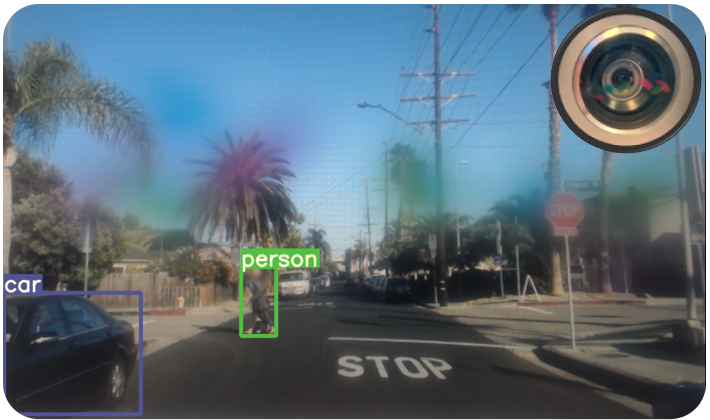}
    \caption{Illustrating the physical translucent patch, placed on the camera lens, which results in the failure to detect the stop sign while correctly identifying the other objects.}
    \label{fig:intro}
\end{figure}

Adversarial attacks have become a major focus of the machine learning research community, primarily in the computer vision domain~\cite{suciu2018does,shafahi2018poison}.
Recent studies have demonstrated the ability to implement evasion attacks (i.e., misclassification of an object) by applying a physical patch on the targeted object.
Some examples include hiding a person using a  small printed cardboard plate~\cite{thys2019fooling} or wearable clothes~\cite{wu2019making, xu2019adversarial, huang2020universal}, concealing a stop sign by attaching small black and white stickers to it~\cite{eykholt2018robust}, or crafting a new stop sign with specific patterns on its background~\cite{chen2018shapeshifter}.
However, the main limitation of those physical attacks is that they require access to the object itself (to apply the adversarial patch). 
Furthermore, to hide multiple objects, an adversarial patch must be applied to each object.

In this paper, we propose a novel physical attack aimed at fooling common object detection models by mounting a carefully crafted patch on the camera lens.
Our attack (patch) is calculated by a gradient-based optimization process which results in a universal adversarial patch that takes the following goals into consideration: 1) successfully hide \textit{all} instances of a target class from the object detector, 2) minimize the attack's impact on the detection of untargeted classes, and 3) produce a printable patch that is as unnoticeable as possible.

To the best of our knowledge, the only study that attempted to implement a camera-based physical attack is the work performed by Li \etal~\cite{li2019adversarial}, in which an image classification model was targeted.
Unlike an image classification model, which predicts the class of a single (dominant) object in the image, an object detection model is capable of \textit{i)} detecting and classifying multiple objects regardless of their location and dominance within the image, and \textit{ii)} processing thousands of candidate bounding boxes centered on each output pixel.
Thus, in our attack all of the candidate objects must be manipulated, significantly complicating the attack.
Furthermore, in this research we study the effect of the physical patch on the untargeted classes, which was not addressed by Li~\etal.

To explore the feasibility of our approach, we first demonstrated the ability to deceive Tesla's advanced driving assistance system (ADAS) by applying two simple fully colored translucent patches on the camera's lens.
We show that the ADAS misclassifies a stop sign (when a red-colored patch is used) and interprets a red traffic light as a green one (when using a cyan-colored patch).
Then, we evaluated our proposed attack on CNN-based object detection models using datasets related to the autonomous car use case.
Since autonomous cars operate in a real-world environment, we use the latest real-time object detection model, YOLOv5~\cite{glenn_jocher_2020_3983579}, a recent improvement to YOLOv3~\cite{redmon2018yolov3}.
In our evaluation, we select the \textit{stop sign} class as the targeted class, with the aim of preventing any of the instances from being detected by the object detector, both in the digital and physical domain.
The results show that we are able to decrease the average precision of the detector for the stop sign class by 42.47\% (digitally) and 42.27\% (physically), while the detection of other object classes remains high.

\noindent We summarize the contributions of our work as follows:
\begin{itemize}[noitemsep]
    \item We present the first camera-based physical adversarial attack on object detection models.
    \item We craft a universal perturbation to fool the model for \textit{all} instances of a specific object class, while maintaining the detection of untargeted objects.
    \item We demonstrate the transferability of the attack when the patch is generated using a surrogate model and then applied to a different model.
    \item The design and optimization process we propose take real-world constraints into account, including printing limitations and accurate patch placement, resulting in a practical attack.
\end{itemize}

%% file: sections/related_work.tex
\section{\label{sec:related}Related Work}

Previous works presenting adversarial attacks in the computer vision domain can be categorized by four main attributes:
\textit{i) model task:} image classification~\cite{krizhevsky2012imagenet}, object detection~\cite{szegedy2013deep}, or image segmentation~\cite{chen2014semantic}; 
\textit{ii) attack type:} digitally generated perturbation or physical perturbations applied in the real-world (either by perturbing the physical object itself~\cite{eykholt2018robust} or by perturbing the sensor's perception of the object~\cite{li2019adversarial}); 
\textit{iii) attacker's knowledge:} full knowledge (white-box)~\cite{su2019one}, or no knowledge (black-box) about the model~\cite{narodytska2016simple};
and \textit{iv) perturbation type:} sparse noise around the entire image~\cite{szegedy2013intriguing} or a centralized dense perturbation in a specific location~\cite{liu2018dpatch}.

As the main novel differentiator of our attack is its physical character, in the following review of related work we categorize the studies based on the \textit{attack type}.

\subsection{Digital Attacks}
When attacks on DNNs in the computer vision domain were first introduced, they mainly targeted CNN-based classification models~\cite{szegedy2013intriguing, goodfellow2014explaining}. 
These kinds of attacks have been extensively studied over the years in research proposing various ways to fool classifiers~\cite{su2019one, moosavi2016deepfool}.
Whereas these methods generate perturbations to fool DNNs on a single and specific image classes, Moosavi-Dezfooli~\etal~\cite{moosavi2017universal} proposed universal adversarial perturbations that can fool any image.
Later, attacks against more complex computer vision tasks were shown. 
For example, Metzen~\etal~\cite{hendrik2017universal} demonstrated that imperceptible perturbations could also fool segmentation models.
However, all of these studies digitally tampered with the input provided to the model. 
Since these attacks lacked significant real-world constraints, they do not naturally transfer to the physical world.

\subsection{Physical Attacks}
In recent years, physical attacks against image classification and object detection models have emerged. 
Kurakin~\etal~\cite{kurakin2016adversarial} took photos of printed adversarial images with a cell phone camera, successfully fooling a pretrained image classifier. 
Eykholt~\etal~\cite{eykholt2018robust} proposed a type of centralized physical perturbation (i.e., applying black and white stickers on stop signs) to fool image classifiers.
Chen~\etal~\cite{chen2018shapeshifter} printed adversarial stop signs by adding specific background patterns, evading the Faster R-CNN~\cite{ren2015faster} object detector, and Sitawarin~\etal~\cite{sitawarin2018darts} crafted toxic signs, similar to original traffic signs, to deceive autonomous car systems.
To avoid facial recognition systems, Sharif~\etal~\cite{sharif2016accessorize} suggested wearing adversarial eyeglass frames.
This work also introduced the \textit{non-printability score}, which forces the adversarial perturbations to use printable colors.

Recently, attacks against person detectors have emerged. 
First, Thys~\etal~\cite{thys2019fooling} printed an adversarial patch on a small cardboard plate, successfully evading YOLOv2~\cite{redmon2017yolo9000} detection.
Following this work, other studies presented attacks in which adversarial patterns were printed on t-shirts~\cite{wu2019making, xu2019adversarial, huang2020universal}.
While these methods create less realistic distortions, Duan~\etal~\cite{duan2020adversarial} used natural patterns that appear legitimate to human observers and applied them on different objects to fool image classifiers.

Unlike the studies mentioned above, in which the perturbation was applied on the targeted object, in our work, we place the perturbation on the sensor collecting the image stream.
While Li~\etal~\cite{li2019adversarial} also applied a physical adversarial perturbation on the camera's lens, their study focused on image classifiers, differing from our work, in which we attack object detectors, a more complex mechanism. 
In addition, we examine the effect of our attack on target class while not reducing the detection of non-targeted classes, which was not an aspect addressed by Li~\etal.

%% file: sections/method.tex
\section{\label{sec:method}Method}

\begin{figure}[]
    \centering
    \begin{subfigure}{0.49\linewidth}
        \centering
        \fbox{\includegraphics[width=0.94\linewidth,trim={0 3cm 0 2.5cm},clip]{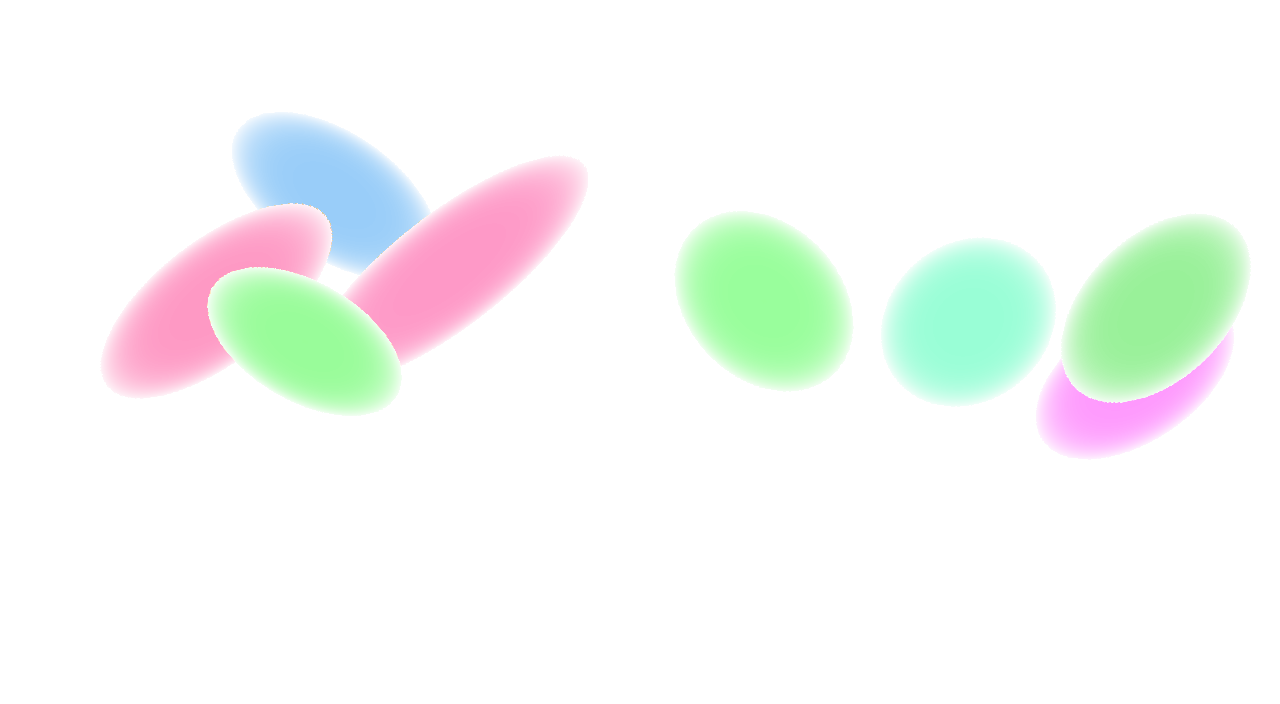}}
        \caption{}
        \label{fig:sub:digital_patch}
    \end{subfigure}
    \begin{subfigure}{.49\linewidth}
        \centering
        \includegraphics[width=0.95\linewidth,trim={0 1.5cm 0 1cm},clip]{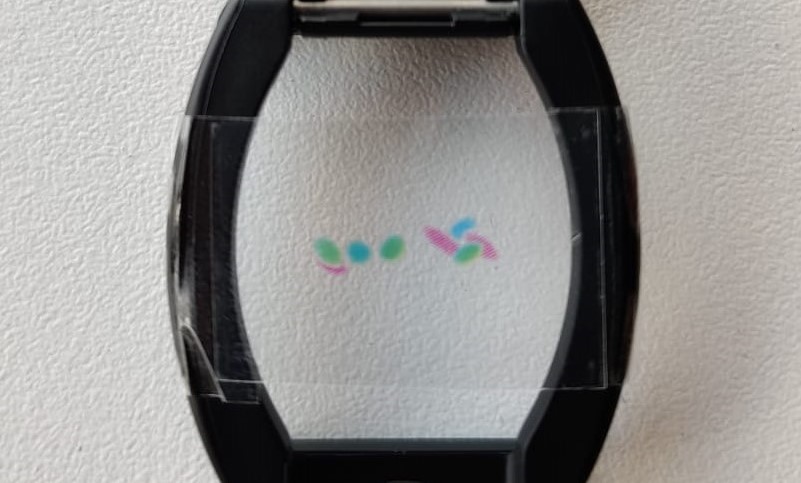}
        \caption{}
        \label{fig:sub:patch_zoom}
    \end{subfigure}
   \begin{subfigure}{.49\linewidth}
        \centering
        \includegraphics[width=0.97\linewidth,trim={0 3.5cm 0 1},clip]{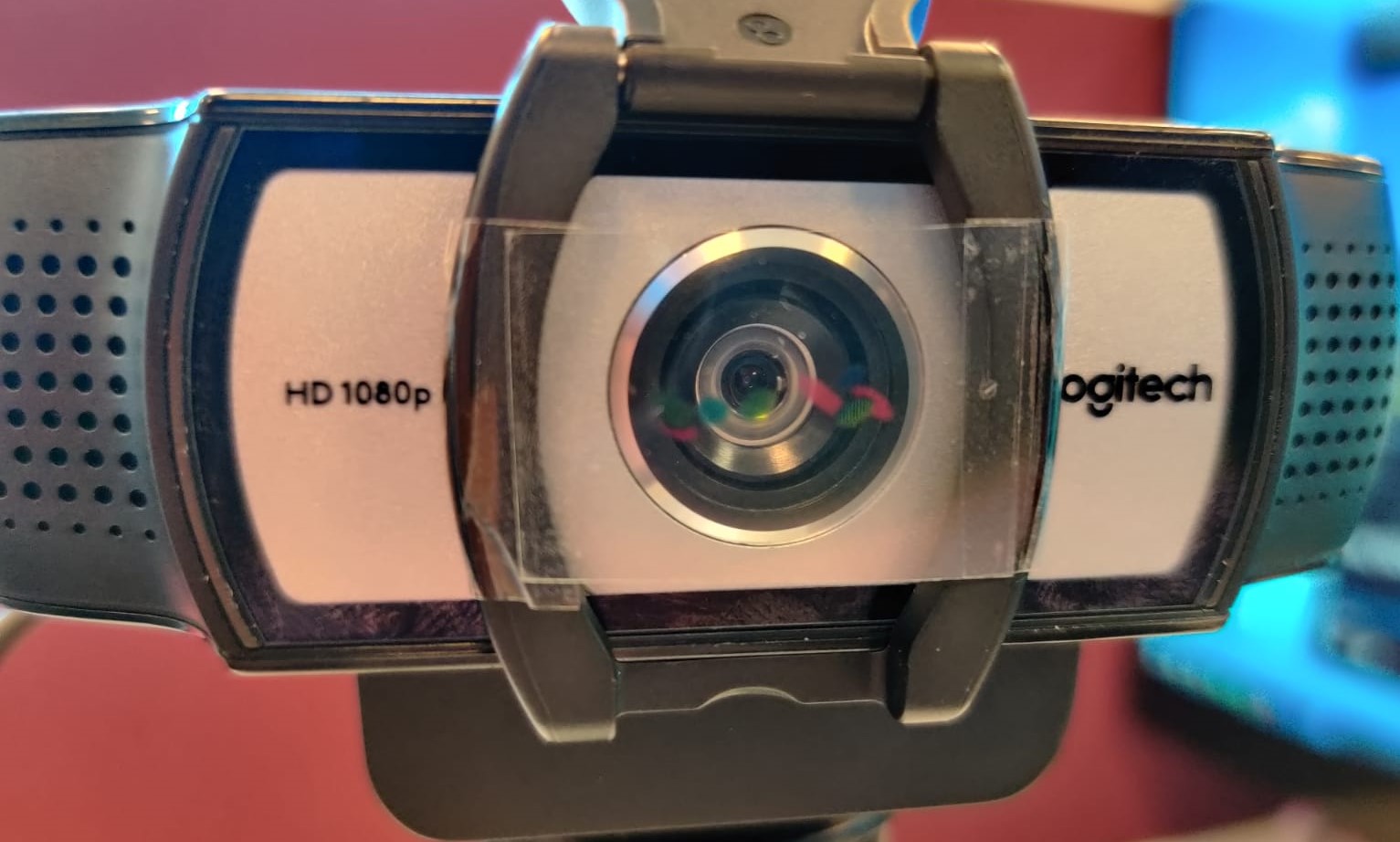}
        \caption{}
        \label{fig:sub:camera_with_patch}
    \end{subfigure}
    \begin{subfigure}{.49\linewidth}
        \centering
        \includegraphics[width=0.97\linewidth,trim={0 13cm 0 0},clip]{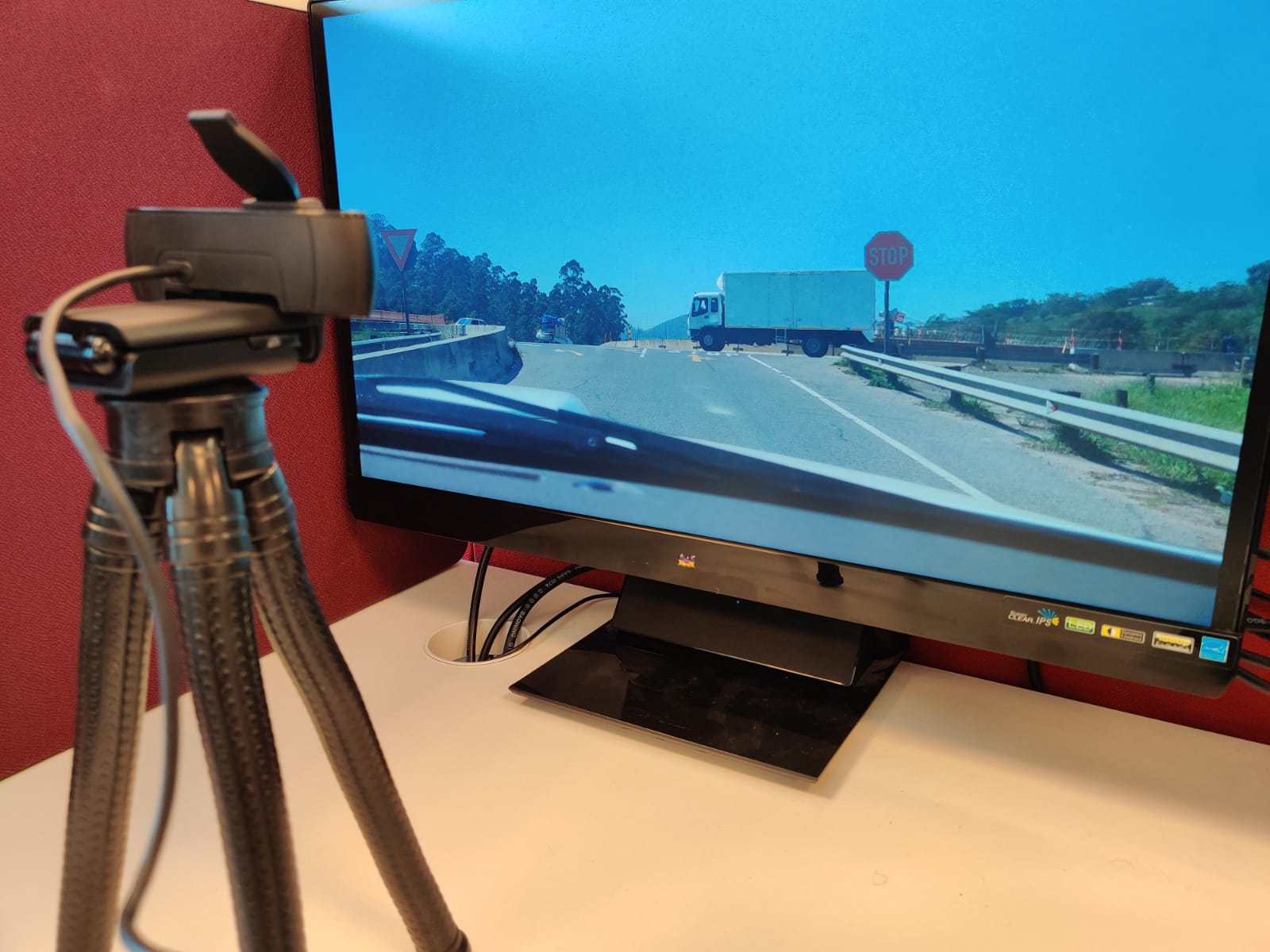}
        \caption{}
        \label{fig:sub:camera_with_screen}
    \end{subfigure}
    \caption{An illustration of: (a) a digital patch with eight shapes; (b) a patch printed on transparent paper; (c) a physical patch applied to the camera's lens; and (d) lab setup of a camera and a screen.}
    \label{fig:physical}
    \vspace{-0.3cm}
\end{figure}

In this research, we aim to generate a printable, translucent universal adversarial perturbation (UAP) that can be used to hide all instances of the selected target class by applying the UAP to the camera's lens.
As also noted by Li~\etal~\cite{li2019adversarial}, accurately attaching pixel-level perturbations onto the camera's lens is impossible and not practical for our case. 
Therefore, we design and craft a \textit{region-level perturbation}, containing several oval shapes printed on transparent paper, to create a practical attack.
By optimizing a custom \textit{loss function}, which considers the main attack goals (i.e., hiding the selected target class instances while maintaining the detection level of the other classes), we generate the region-level perturbation.
We assume that the attacker has direct access to the camera's lens to apply the patch (e.g., via a supply chain).

In this section, we start by describing the design requirements for the patch (Section~\ref{sec:patch_design}).
Then, we present the process of optimizing the parameters of the patch in order to achieve a successful attack (Section~\ref{sec:patch_optimization}).

\begin{figure*}[]
\centering
    \includegraphics[width=0.95\linewidth]{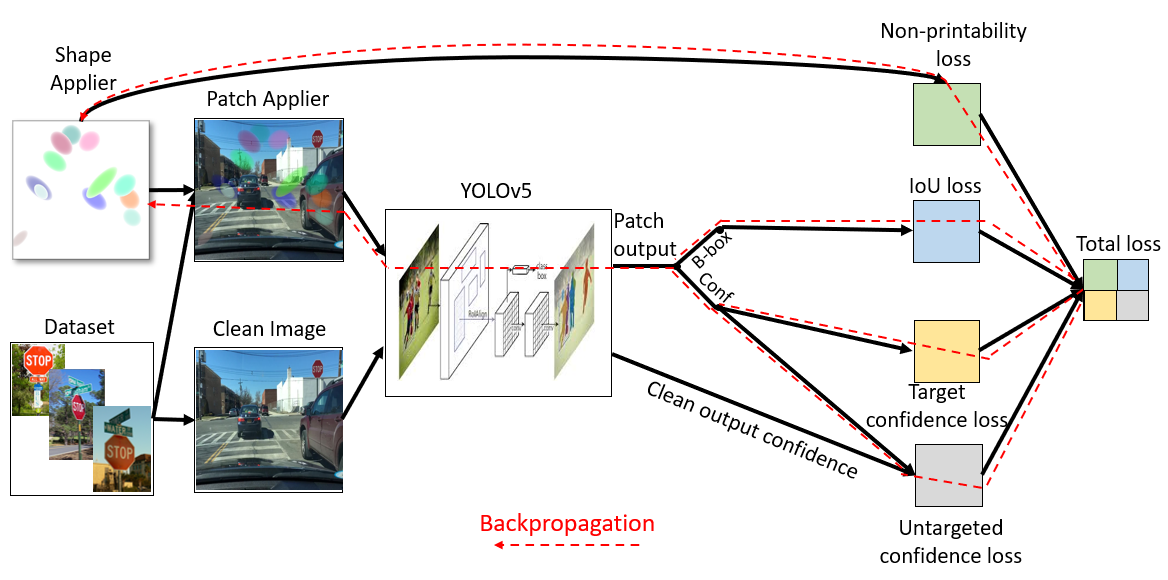}
    \caption{Overview of our method's pipeline.}
    \label{fig:pipeline}
    \vspace{-0.5cm}
\end{figure*}

\subsection{\label{sec:patch_design}Patch Design}

\noindent \textbf{From digital to physical setup.} 
Our adversarial perturbation is computed digitally and then applied as a real physical patch.
Therefore, when computing the digital patch, we need to consider the effect of printing and applying a translucent physical patch on the camera's lens.
As Li~\etal~\cite{li2019adversarial} suggested, an approximation of that effect can be achieved using an alpha blending process between the original image and the digital patch.
Therefore, in this study, we define our patch as a 2D image with four channels. The first three channels represent RGB colors supplemented with a fourth \textit{alpha} channel representing how opaque each pixel is.
The result of performing alpha blending between the \textit{original} image and a translucent patch at pixel $(i,j)$ is defined as follows:
\begin{multline}
    \label{eq:alpha_blending}
    perturbed(i,j) = original(i,j) * (1-\alpha(i,j))+ \\ \gamma{(i,j)}*\alpha(i,j)
\end{multline}
\noindent where $perturbed$ is the resulting image consisting of just RGB channels, $\alpha$ represents the patch's alpha channel, and $\gamma$ represents RGB triplets.

\vspace{0.1cm}
\noindent \textbf{Patch structure.}
As illustrated in Figure~\ref{fig:sub:digital_patch}, our patch comprises $n$ (configurable parameter) blurry oval shapes; these shapes are initialized as fully colored dots with shearing in the $x$ and $y$ directions.
The following attributes define a single shape:
\begin{itemize}
    \item $(x_c, y_c)\in[-1,1]\subset\mathbb{R}^2$ - a tuple representing a shape's center with regard to the center of the image; for example, a shape centered at $(0,0)$ will be placed in the center of the image.
    
    \item $r\in[r_{min},r_{max}]\subset\mathbb{R}$ - the shape's radius in relation to the patch size.
    
    \item $(sh_x, sh_y)\in[-1,1]\subset\mathbb{R}^2$ - a tuple representing the shape's shear in the $x$ and $y$ directions.
    
    \item $\gamma\in\mathbb{R}^3$ - a triplet representing an RGB color.
    
    \item $\alpha\in[0,1]\subset\mathbb{R}$ - represents the alpha channel. 
\end{itemize}

\vspace{0.1cm}
\noindent \textbf{Patch blending.}
Usually, when crafting a pixel-level patch, the \emph{total variance} factor~\cite{sharif2016accessorize} is also included in the custom loss function to ensure that the optimizer favors smooth color transitions between adjacent pixels, thus preventing noisy images.
Since we aim to generate a region-level patch, the configuration of the patch's alpha channel is implemented as a predefined total variation constraint, forcing neighboring pixels within the same shape to be similar.
We achieve this by defining a position-dependent alpha channel at pixel $(i,j)$ for a single shape as follows:
\begin{equation}
    \begin{gathered}
        \alpha(i,j) = a_{max}*(-s*d(i,j)^{\beta}+1)\\
        d(i,j) = \frac{(i - x_{c,norm})^2+(j - y_{c,norm})^2}{r_{norm}^2}\\
        x_{c,norm} = (1-x_c)*\floor{\frac{p_w}{2}}\\
        y_{c,norm} = (1-y_c)*\floor{\frac{p_h}{2}}\\
        r_{norm} = r * \floor{\frac{min(p_w, p_h)}{2}}
    \end{gathered}
    \label{eq:alpha}
\end{equation}
where the parameters above represent the following:
\begin{itemize}
    \item $a_{max}\in[0,1]\subset\mathbb{R}$ - the maximum value of the alpha channel.
    \item $s\in[0,1]\subset\mathbb{R}$ - controls the minimum value of the alpha channel, where $\alpha_{min}=\alpha_{max}*(1-s)$.
    \item $\beta\in\mathbb{R_+}$ - exponential drop-off.
    \item $(p_w,p_h)\in\mathbb{N}^2$ - patch width and height.
    \item $d(i, j)\in[0,1]\subset\mathbb{R}$ - the normalized distance of pixel $(i, j)$ from the shape's normalized center $(x_{c,norm},\,y_{c,norm})$.
\end{itemize}

\noindent Equation~\ref{eq:alpha} produces distance-dependent opacity along the shapes' area, forming a smooth shape. 
Pixels close to the shape's center ($d\to0$) will have an alpha value closer to $\alpha_{max}$, whereas pixels that lie close to the shape's edge ($d\to1$) will have near $\alpha_{min}$ value. 
The drop-off parameter $\beta$ controls the smoothness intensity.

\vspace{0.1cm}
\noindent \textbf{Shape's positioning and shearing.}
To control the positioning and shearing of each shape on the patch, we use 2D affine transformations~\cite{stearns1995method}, characterized by the following affine matrix:
\begin{equation}
    \begin{bmatrix}
        1 & sh_x & x_c\\
        sh_y & 1 & y_c
    \end{bmatrix}
\end{equation}

\noindent Intuitively, the shape's location is characterized by discrete pixel coordinates.
However, since we use a gradient-based optimization process, discrete parameters cannot be used.
Therefore, we use affine transformation to represent a continuous and differentiable form of the shape's location.

\vspace{0.1cm}
\noindent \textbf{Attack parameters.}
The parameters discussed in this section can be divided into two groups:
\begin{itemize}
    \item a set of manually chosen (input) parameters: number of shapes $n$ and $\Theta_{manual}$ which characterizes all of the shapes on our patch:
    \begin{equation}
        \Theta_{manual} = (\alpha_{max}, s, \beta, (r_{min}, r_{max}))
    \end{equation}
    The specific values chosen for $\Theta_{manual}$ are essential for limiting the amount of noise applied to our patch and simulating a printed physical patch, which will be further explained in Section~\ref{sec:eval}.
    \item The free parameters $\Theta_{free}$ are optimized by the proposed algorithm:
    \begin{equation}
        \begin{gathered}
            \theta = ((x_c, y_c), r, (sh_x, sh_y), \gamma)\\
            \Theta_{free} = \{\theta_1,..,\theta_n\} 
        \end{gathered}
    \end{equation}
    where $\theta$ characterizes a single shape and $\Theta_{free}$ is a composition of all of the shape's parameters.
\end{itemize}

\subsection{\label{sec:patch_optimization}Patch Optimization}

To optimize $\Theta_{free}$, we compose a custom loss function consisting of four components:
\begin{equation}
    \label{eq:loss_total}
    \begin{gathered}
        \ell_{total} = w_1 * \ell_{target\,conf} + w_2 * \ell_{IoU}\\
        + w_3 * \ell_{untargeted\,conf} + w_4 * \ell_{nps}\\
        \sum_i w_i=1
    \end{gathered}
\end{equation}

\noindent
The optimization process minimizes $\ell_{total}$, aiming to achieve three main goals, each of which will be discussed throughout this section, noting their relation to the components presented in Equation~\ref{eq:loss_total}.
To determine the optimal weight values $w_i$, we use the grid search approach. We also allow zeroing $w_1$ and $w_2$ during the hyperparameter search to study whether they are both necessary to achieve our goals.

Since our entire attack is differentiable, we use an automatic differentiation tool kit (PyTorch) to optimize our patch parameters using the backpropagation algorithm, as shown in Figure~\ref{fig:pipeline}.

\vspace{0.1cm}
\noindent \textbf{Eliminating the detection of the target class.}
The inference output of object detection models provides two unique defining details for each detected object: the bounding box and confidence score.
We use these details to achieve the following goals:
\begin{itemize}
    \item Minimize the confidence score of the target class:
    \begin{equation}
        \ell_{target\,conf} = Pr(objectness)* Pr(target\,class)
    \end{equation}
    where $Pr(objectness)$ and $Pr(class)$ correspond to the YOLO output, which consists of two confidence scores for each cell in the final detection grid: (a) the objectness score - whether a specific cell in the grid contains any object, (b) the class score -  whether a specific cell in the grid contains a specific class.
    The model outputs a detection only if the confidence score surpasses a certain threshold.
    As noted by Thys~\etal~\cite{thys2019fooling}, it is possible to minimize $Pr(objectness)$ and $Pr(class)$ individually.
    However, in our preliminary experiments, the only combination that led to adequate results was the product of these components.
    
    \item Minimize the Intersection over Union (IoU) score between the predicted bounding box and the ground truth bounding box of the target class:
    \begin{equation}
        \ell_{IoU} = IoU^{ground\,truth}_{predicted}(target\,class)
    \end{equation}
    By doing this, we steer the shapes in our patch toward learning better positions, causing the detector to incorrectly predict the location of bounding boxes and resulting in the misdetection of the object's location.
	Although our attack aims to hide a selected target class, incorrectly placing a bounding box can also negatively affect the detector's performance.
    When the location of the predicted bounding box matches that of the ground truth bounding box with high accuracy, the penalty of this component is greater.
\end{itemize}

\vspace{0.1cm}
\noindent \textbf{Maintaining the detection of untargeted classes.}
Since our patch is not placed on the target object, we address the issue of affecting the detection of the untargeted classes:
\begin{multline}
\label{loss_untargeted}
    \ell_{untargeted\,conf}=\\
    \frac{1}{M}
        \sum\limits_{\substack
        {cls\in image \\ cls \neq target}}{|conf(cls,clean)-conf(cls,patch)|}
\end{multline}
where $conf(cls,image\,type)$ represents the confidence score of class $cls$ in image $image\,type$, as explained earlier in this section, and $M$ represents the number of classes in the image.

When untargeted classes are detected correctly, the loss will be closer to zero. On the other hand, when the detector does not detect untargeted classes, the penalty of this component is greater.

\vspace{0.1cm}
\noindent \textbf{Generating a printable patch.}
We include the non-printability score (NPS)~\cite{sharif2016accessorize}, which represents how closely digital colors match colors printed by a standard printer.
\begin{equation}
\label{loss_nps}
    \ell_{nps}=\sum\limits_{c_{patch}\in P}\min\limits_{c_{print}\in C}{|c_{patch} - c_{print}|}
\end{equation}
where $c_{patch}$ is a color in our patch \textit{P} (i.e., shape color) and $c_{print}$ is a color in the set of printable colors \textit{C}.
This loss penalizes colors that are far from the set of printable colors.

%% file: sections/tesla.tex
\section{Motivating Example: Tesla Use Case}

\begin{figure}[t!]
    \centering
    \includegraphics[width=0.85\linewidth]{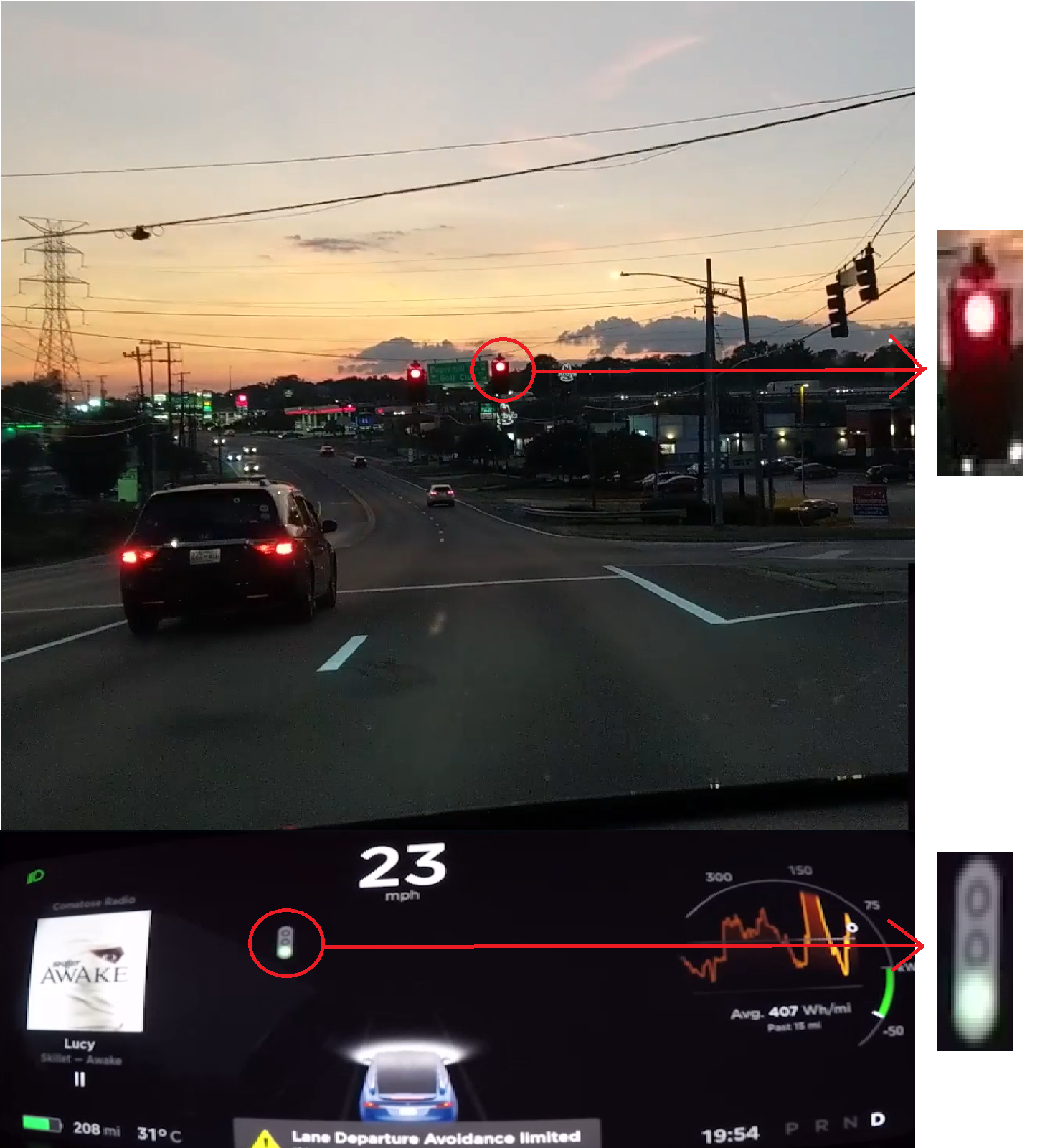}
    \caption{Camera-based attack on a Tesla Model X, using a cyan-colored patch. 
    The street view in the upper part of the image shows the actual scene with a red traffic light. 
    The lower part of the image shows that the car's navigation screen interprets the traffic light's color as green.}
    \label{fig:tesla-tl}
    \vspace{-0.4cm}
\end{figure}

As a proof of concept, we demonstrated the ability to successfully perform a camera-based attack on an object detection model by applying a physical on-sensor patch.
After applying fully colored patches (i.e., not generated by our attack) on the front camera of a Tesla Model X, we were able to deceive the car's advanced driving assistance system (ADAS):
\begin{itemize}[noitemsep]
    \item Traffic light attack - by applying a cyan-colored patch, we managed to change the ADAS's perception of a traffic light's color, so that it interpreted a red light as a green light, as shown in Figure~\ref{fig:tesla-tl}.
    \item Stop sign attack - by applying a red-colored patch, we were able to hide a stop sign and prevent it from being detected.
\end{itemize}
A demo of both experiments can be found here: \url{https://youtu.be/IO1P2KDsWVE}.

%% file: sections/evaluation.tex
\section{\label{sec:eval}Evaluation}

\input{sections/physical_examples}

We evaluate our attack on the use case of autonomous cars, in which object detection models are used to identify obstacles and road signs.
Specifically, we aim to devise an attack that will cause the object detection model to fail to detect stop signs while maintaining its capability of detecting other object classes.

We first experiment in the digital domain by using alpha blending between the original images and several different types of patches (see Equation~\ref{eq:alpha_blending}).
Then, we evaluate the performance of the patches in a real-world setup by printing and attaching them to a camera lens, as shown in Figure~\ref{fig:physical}.

\vspace{0.05cm}
\noindent \textbf{Models.}
We evaluated our attack using the YOLOv5 one-stage detector~\cite{glenn_jocher_2020_3983579} (an improvement to YOLOv3~\cite{redmon2018yolov3}) in a white-box setting.
In addition, we examined our patch's transferability to other detectors, YOLOv2~\cite{redmon2017yolo9000} and Faster R-CNN~\cite{ren2015faster}, as a black-box setting.
For all of the detectors we use pretrained weights on the MS-COCO~\cite{lin2014microsoft} dataset.
MS-COCO contains 80 object categories from several domains. 
We selected eight relevant categories: person, bicycle, car, bus, truck, traffic light, fire hydrant, and stop sign.

\vspace{0.1cm}
\noindent \textbf{Datasets.}
We use a combination of multiple driving-related datasets, from which we extracted images containing stop signs, to improve the robustness of our patch:
\begin{itemize}
    \item LISA traffic sign dataset~\cite{mogelmose2012vision} - tens of videos split into frames containing U.S. traffic signs; $\sim$500 images containing stop signs.
    \item Mapillary Traffic Sign Dataset (MTSD)~\cite{ertler2019mapillary} - a diverse street-level dataset obtained from a rich geographic area; $\sim$750 images containing stop signs.
    \item Berkeley DeepDrive (BDD)~\cite{yu2020bdd100k} - videos of the driving experience covering many different times of the day, weather conditions, and driving scenarios; $\sim$500 images containing stop signs.
\end{itemize}
Hence, the full dataset contains $\sim$1750 images of stop signs.
Since the LISA and MTSD datasets only have annotations of traffic signs, we use our object detection model (YOLOv5) to annotate (label) the rest of the classes examined in this paper.
These annotations are later used as the ground truth.

\begin{figure}[t!]
    \begin{subfigure}{1.\linewidth}
        \centering
        \includegraphics[width=0.9\linewidth,trim={0.6cm 0.2cm 0.6cm 0.6cm},clip]{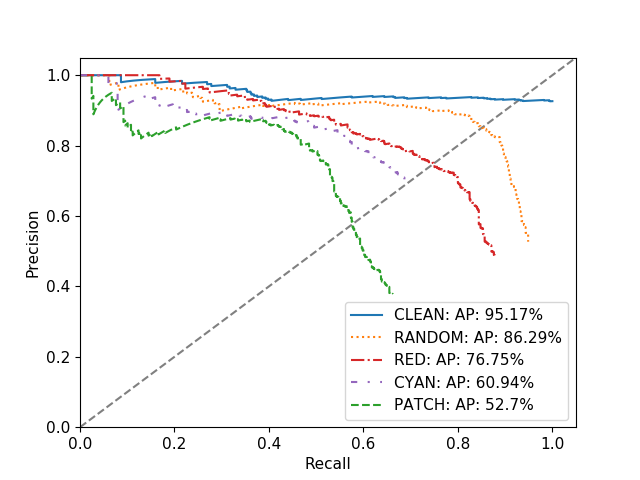}
        \caption{Stop sign class.}
        \label{fig:target-pr}
    \end{subfigure}
    \begin{subfigure}{1.\linewidth}
    \centering
       \includegraphics[width=0.9\linewidth,trim={0.6cm 0.2cm 0.6cm 0.6cm},clip]{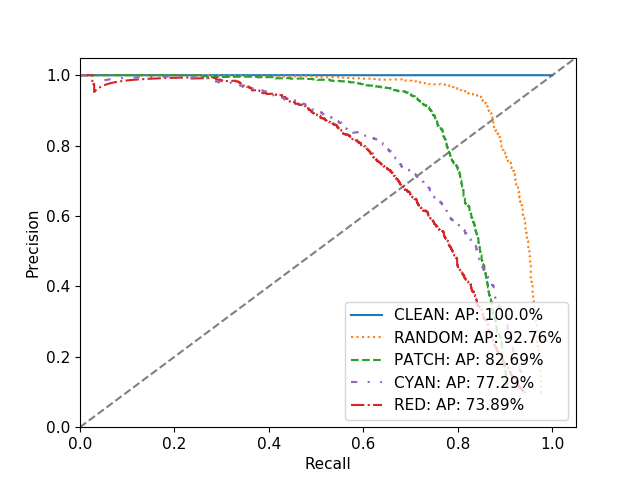}
        \caption{Other (untargeted) classes.}
        \label{fig:other-pr}
    \end{subfigure}
    \caption{Precision-Recall (PR) curves of our method (PATCH) with eight shapes compared to different patch types (RANDOM, RED, CYAN) and the original images (CLEAN) on the YOLOv5 (white-box setting).}
    \vspace{-0.4cm}
\end{figure}

\vspace{0.1cm}
\noindent \textbf{Evaluation setup.}
We split our full dataset into training, validation, and test sets.
The training and validation sets consist of images from the BDD and MTSD datasets (randomly chosen with a split ratio of 90\% and 10\%, respectively), while the test set contains images from the LISA dataset.
By dividing the full dataset this way, we achieve two goals: (1)
there is no correlation between the training/validation sets and the test set; thus, achieving good results on the test set implies that our patch is unbiased and universal, and (2) since the LISA dataset is comprised of multiple videos split into frames, it allows us to demonstrate our patch's effectiveness in the physical domain by testing it on driving videos.

\vspace{0.1cm}
\noindent \textbf{Evaluation metrics.}
We use the average precision (AP) for digital attacks, representing the area under the precision-recall (PR) curve for both the selected target class and untargeted classes.

In addition, to quantify our physical patch's success in the real world, we define the following metric:
\begin{equation}
    Fooling\,Rate(class) = \frac{\#\,fooled\,objects\,(class)}{\#\,total\,objects\,(class)}
\end{equation}
where an object of category \textit{class} is considered `fooled' if it was not output by the detector (the confidence score does not surpass the minimum threshold). 
We set our minimum confidence score threshold at 0.4, a point where our detector achieved an AP of 95.17\% on the original images containing the stop sign class in the digital domain.

\vspace{0.1cm}
\noindent \textbf{Types of patches evaluated.}
We compare our patch's performance to several different translucent patches: (a) CLEAN - the original image without a patch, (b) PATCH - our optimized patch, (c) RANDOM - a baseline patch with the same number of shapes as PATCH and random initialization of our attack's optimized parameters $\Theta_{free}$: location, color, and shearing, (d) RED - a fully red-colored patch, (e) CYAN - a fully cyan-colored patch. 
The different types of patches evaluated are presented in Figure~\ref{fig:attacks}.

\noindent \textbf{Implementation details.}
Throughout this section the results presented are obtained using the following manual parameters:
1) $s=0.9$, to achieve near zero values at the shape's edge, for smooth transitions between pixels inside and outside the shape's area; 2) $\beta=2.5$, to apply high intensity around the shape's center, which will be caused in the physical patch by the printer; and 3) the upper radius bound $r_{max}=0.25$, to bind the number of pixels changed in the perturbed image, while the lower bound $r_{min}=0.03$ is the minimum printable shape.

The settings of our free parameters $\Theta_{free}$ are randomly initialized and updated using the Adam optimizer~\cite{kingma2014adam}. 
The initial learning rate is set at $5e^{-3}$, except for the radius update, which is initially set at $8e^{-4}$.
Moreover, using a grid search algorithm, we found the optimal $w_i$ setting to be $w_1 = 0.74\,,w_2=0.15\,,w_3=0.1\,,w_4=0.01$

\subsection{Digital Attacks}
To evaluate our patch's effectiveness, we conduct digital experiments in which our optimized patch is applied to images in the test set, using alpha blending, as explained in Equation~\ref{eq:alpha_blending}.

\noindent \textbf{White-box attack performance.}
First, we examine our attack's performance in a \textit{white-box} setting in which our patch is optimized using the weights of the YOLOv5 detector and evaluated on the same model.
As shown in Figure~\ref{fig:target-pr}, using a patch with eight shapes with $\alpha_{max} = 0.4$, we were able to degrade the detector's AP to 52.7\%, which is a decrease of over 42.47\% from the detector's best performance (which is 95.17\%).
In addition, of the patches examined, our patch has the greatest effect on the detector's performance for the \textit{stop sign} class.

Since we do not apply our patch on a specific object but rather on the entire image, it might interfere with the detection of untargeted classes; thus, a comprehensive evaluation must address that issue as well. 
The results in Figure~\ref{fig:other-pr} show that our patch has a minimal effect on untargeted classes. 
The detector was able to achieve an AP of 82.69\%, for the untargeted classes which is only a 10.07\% difference from the random patch and better than simple fully-colored patches.
It should be noted that our detector's AP on untargeted classes is 100\%, because the ground truth labels on the LISA dataset were generated by our detector, as mentioned earlier in this section.
Unlike our patch, which affects the detection of the stop sign class more than other classes, the rest of the attacks have a similar effect on all of the classes, both targeted and untargeted.

\vspace{0.2cm}
\begin{table}[b]
\centering
\small
\begin{tabular}{|c|c|c|} 
\hline
\begin{tabular}[c]{@{}c@{}}\# Shapes ($n$)\end{tabular} & \begin{tabular}[c]{@{}c@{}}`Stop Sign' Class \end{tabular} & \begin{tabular}[c]{@{}c@{}}Other Classes \end{tabular}  \\ 
\hline
3    & 91.15\%  & 95.44\%    \\ 
\hline
5   & 77.45\%     & 91.72\%      \\ 
\hline
7     & 65.01\%   & 83.23\%      \\ 
\hline
10    & 53.11\%   & 77.65\%      \\ 
\hline
15    & 47.9\%  & 72.49\%   \\ 
\hline
\end{tabular}
\vspace{-0.2cm}
\caption{Average precision as a function of $n$}
\label{tab:num_of_shapes}
\vspace{-0.2cm}
\end{table}

\noindent \textbf{Attack's performance for different parameter values.}
A significant aspect of adversarial attacks refers to the amount of noise added to the resulting image, which is mainly controlled by manual parameter setting.
The noise can have two kinds of impact, affecting the deception rate as follows: 
\begin{itemize}
    \item The number of pixels changed - in our attack this is affected by two parameters: the number of shapes $n$ and the shapes' radius $r$.
    Since $r$ is a free parameter, we examine the effect of the number of shapes with $\alpha_{max} = 0.4$.
    As shown in Table~\ref{tab:num_of_shapes}, we can see that the AP decreases as we add more shapes.
    It should be noted that at some point, the AP does not decrease as we add more shapes, because shapes may overlay each other.
    
    \item The patch's dominance in the resulting image - in our attack the dominance is mainly influenced by $\alpha_{max}$; therefore, we use a fixed number of shapes (eight) to study $\alpha_{max}$'s influence.
    As shown in Table~\ref{tab:alpha_max}, we are able to decrease the detector's AP as the patch's $\alpha_{max}$ increases in the resulting image, eliminating a large portion of stop sign instances when setting the maximum opacity of the patch to 90\%. 
    While high opacity achieves a higher fooling rate,
    at some level the patch becomes perceptible to the human eye. 
    Thus, the setting of $\alpha_{max}$ must address the patch's effectiveness on the target class and keep the patch as imperceptible as possible.
\end{itemize}
Moreover, adding too many shapes or setting the value of $\alpha_{max}$ too high largely affects the detection of the untargeted classes, which is something that we want to avoid.

\begin{table}[h]
\centering
\small
\begin{tabular}{|c|c|c|} 
\hline
\begin{tabular}[c]{@{}c@{}}$\alpha_{max}$\end{tabular} & \begin{tabular}[c]{@{}c@{}}`Stop Sign' Class \end{tabular} & \begin{tabular}[c]{@{}c@{}}Other  Classes \end{tabular}  \\ 
\hline
0.1  & 93.85\%  & 98.26\%  \\ 
\hline
0.3  & 70.13\% & 88.25\%    \\ 
\hline
0.5     & 51.75\%    & 81.93\% \\ 
\hline
0.7     & 38.61\%     & 78.76\% \\ 
\hline
0.9 & 36.55\% & 70.45\% \\
\hline
\end{tabular}
\caption{Average precision as a function of the $\alpha_{max}$}
\label{tab:alpha_max}
\vspace{-0.2cm}
\end{table}

\noindent \textbf{Transferability of the attack.}
To perform a comprehensive evaluation of our patch, we further investigate its performance on detectors that it was not trained on (black-box setting), which means that our patch (eight shapes and $\alpha_{max}=0.4$) was optimized using a surrogate model and tested on other models.
In Table~\ref{tab:transferability}, we present our patch's performance on two attacked models: YOLOv2~\cite{redmon2017yolo9000} and Faster R-CNN~\cite{ren2015faster}.
We show that our patch can successfully deceive models it was not trained on, reducing the AP of the YOLOv2 and Faster R-CNN on the stop sign class to 57.36\% and 54.53\% respectively, while maintaining an AP for untargeted classes that is close to the detector's best performance for this task.

\begin{table}[h]
\small
\begin{tabular}{|c|c|c|c|c|}
\hline
             & \multicolumn{2}{c|}{`Stop Sign' Class} & \multicolumn{2}{c|}{Other Classes} \\ \hline
Model/Attack & CLEAN              & PATCH             & CLEAN            & PATCH           \\ \hline
YOLOv5       & 95.17\%            & 52.7\%            & 100\%            & 82.69\%         \\ \hline
YOLOv2       & 81.54\%            & 57.36\%           & 59.13\%          & 54.92\%         \\ \hline
Faster R-CNN & 94.31\%            & 54.53\%           & 78.31\%          & 70.36\%         \\ \hline
\end{tabular}
\caption{Average precision in black-box setting: patch trained on YOLOv5 and evaluated on other object detectors}
\label{tab:transferability}
\vspace{-0.5cm}
\end{table}

\subsection{Physical attacks}
Finally, to evaluate our patch's performance in the real world, we print it on transparent paper (Figure~\ref{fig:sub:patch_zoom}) and place it on the camera's lens (Figure~\ref{fig:sub:camera_with_patch}), filming a computer screen that is projecting videos of our test set (Figure~\ref{fig:sub:camera_with_screen}).
Since the physical attack evaluation requires access to the object detection model's predictions, we used a testbed  which contained the following equipment: a Logitech C930 web camera to simulate the autonomous car's camera, a 21-inch computer screen to project videos on, simulating real driving, and YOLOv5 to simulate the car's ADAS.
To print our patch (0.6x0.33 inches) on transparent paper, we used a Xerox 6605DN laser printer.

We first evaluate the detector's best performance on the test set videos without any attack (i.e., no patch is applied to the camera's lens) and then use this as a reference (ground-truth) to examine the effect of different patches. 
As the results presented in Table~\ref{tab:physical} show, our physical patch was able to cause the detection model to fail to detect 42.27\% of the total number of stop sign instances (compared to 42.47\% in the digital attack), while still detecting nearly 80\% of the untargeted class instances (compared to 82.69\% in the digital attack).
In contrast to the results obtained by our patch, the other patches have major disadvantages: 1) the random patch could not achieve an adequate fooling rate on the stop sign class, and 2) the fully colored patches performed poorly on the untargeted classes (similar to the effect of completely blocking the camera's lens).

During the physical experiments we observed a specific trend when our patch was used - in most of the scenes, the stop sign is detected at a very late stage, which leads to a very small window of time for the ADAS to respond.

\begin{table}[t]
\centering
\small
\begin{tabular}{|c|c|c|c|c|}
\hline
Class/Attack
& \begin{tabular}[c]{@{}c@{}}PATCH\end{tabular} & \begin{tabular}[c]{@{}c@{}}RANDOM\end{tabular} & \begin{tabular}[c]{@{}c@{}}RED\end{tabular} & \begin{tabular}[c]{@{}c@{}}CYAN\end{tabular} \\ \hline
Stop sign  & 42.27\%     & 20.57\%   & 93.3\%     & 98.9\%     \\ \hline
Others        & 21.54\%     & 19.27\%      & 82.7\%  & 81.6\%     \\ \hline
\end{tabular}
\caption{Fooling rate for the stop sign class and other classes for physical patch attacks}
\label{tab:physical}
\vspace{-0.5cm}
\end{table}

%% file: sections/physical_examples.tex
\begin{figure*}[t!]
    \captionsetup[subfigure]{labelformat=empty}
    \centering
    \begin{subfigure}{.19\linewidth}
        \centering
        \caption{CLEAN}
        \includegraphics[width=\linewidth]{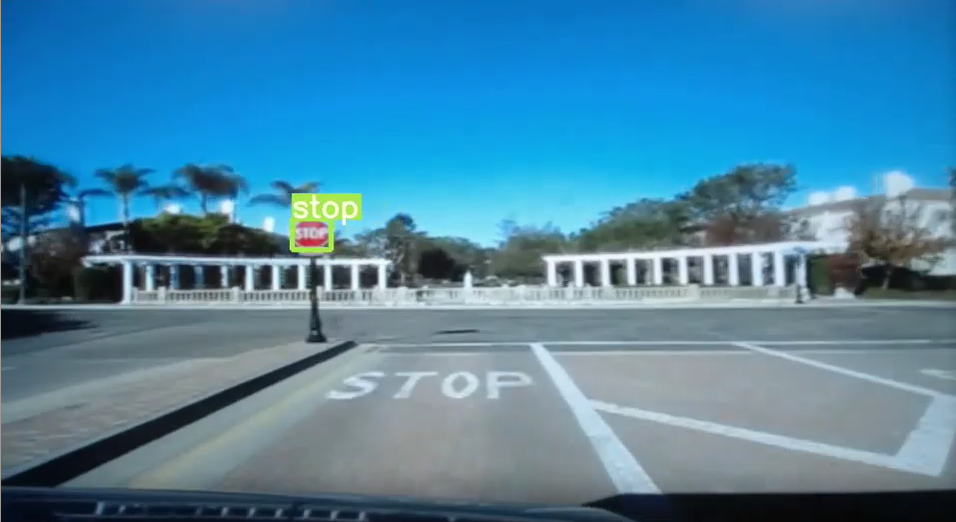}
    \end{subfigure}
    \begin{subfigure}{.19\linewidth}
        \centering
        \caption{PATCH}
        \includegraphics[width=\linewidth]{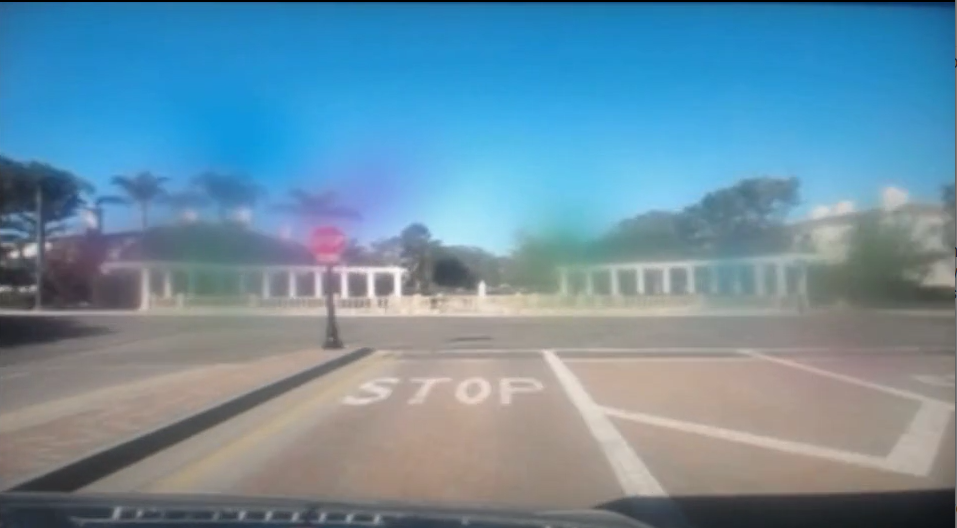}
    \end{subfigure}
    \begin{subfigure}{.19\linewidth}
        \centering
        \caption{RANDOM}
        \includegraphics[width=\linewidth]{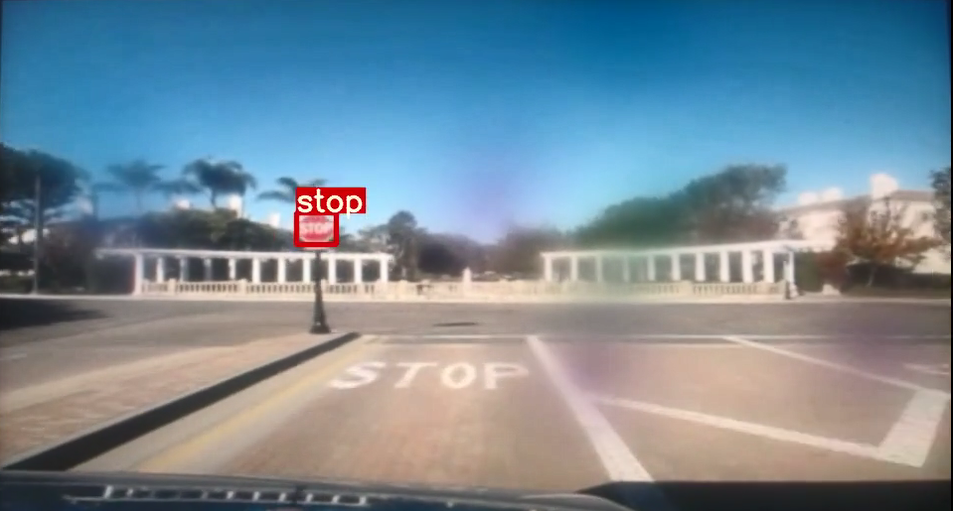}
    \end{subfigure}
    \begin{subfigure}{.19\linewidth}
        \centering
        \caption{RED}
        \includegraphics[width=\linewidth]{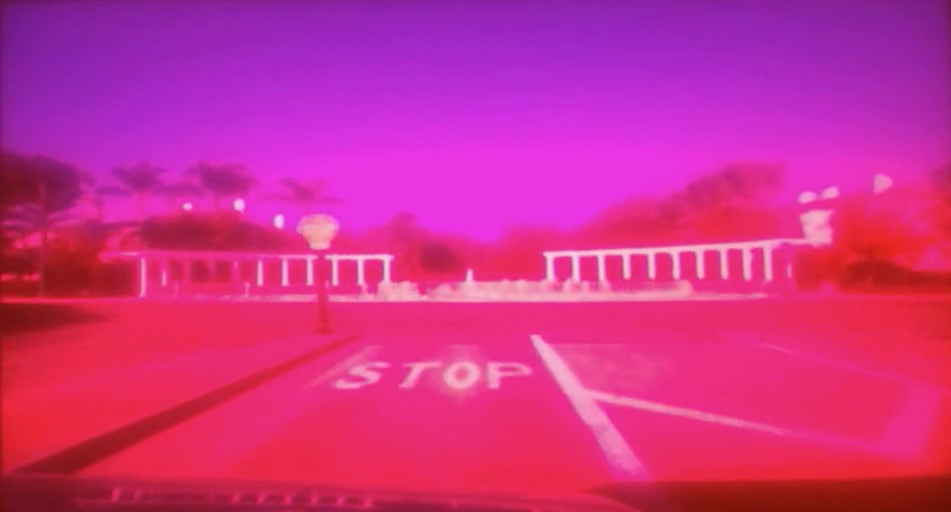}
    \end{subfigure}
    \begin{subfigure}{.19\linewidth}
        \centering
        \caption{CYAN}
        \includegraphics[width=\linewidth]{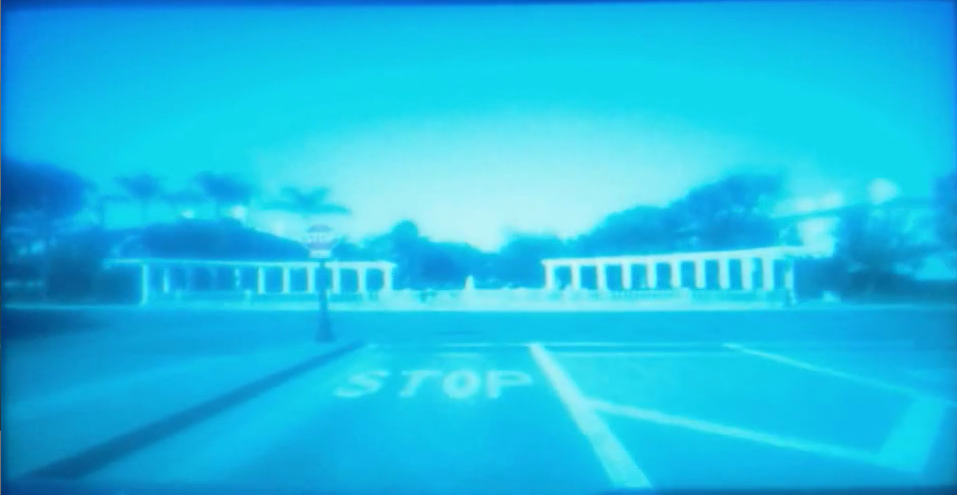}
    \end{subfigure}
    \begin{subfigure}{.19\linewidth}
        \centering
        \includegraphics[width=\linewidth]{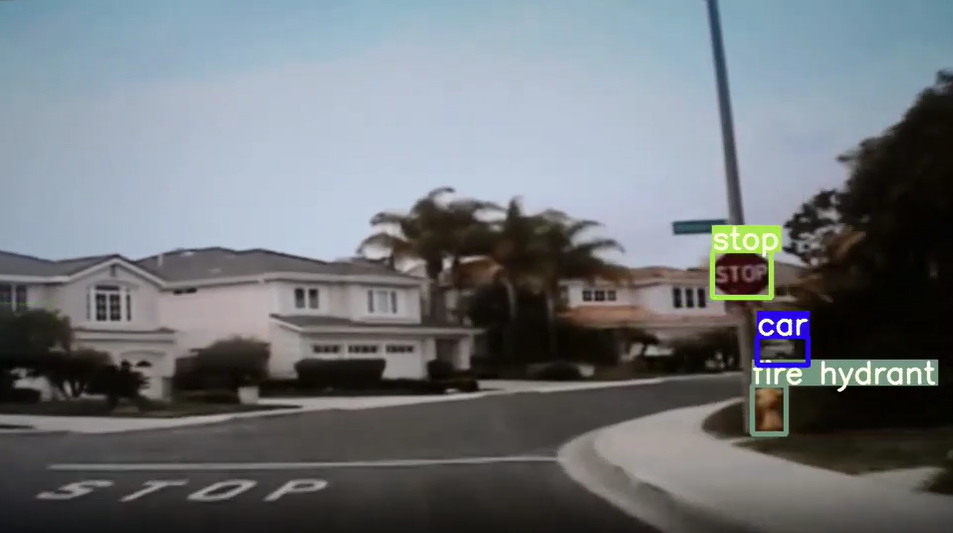}
    \end{subfigure}
    \begin{subfigure}{.19\linewidth}
        \centering
        \includegraphics[width=\linewidth]{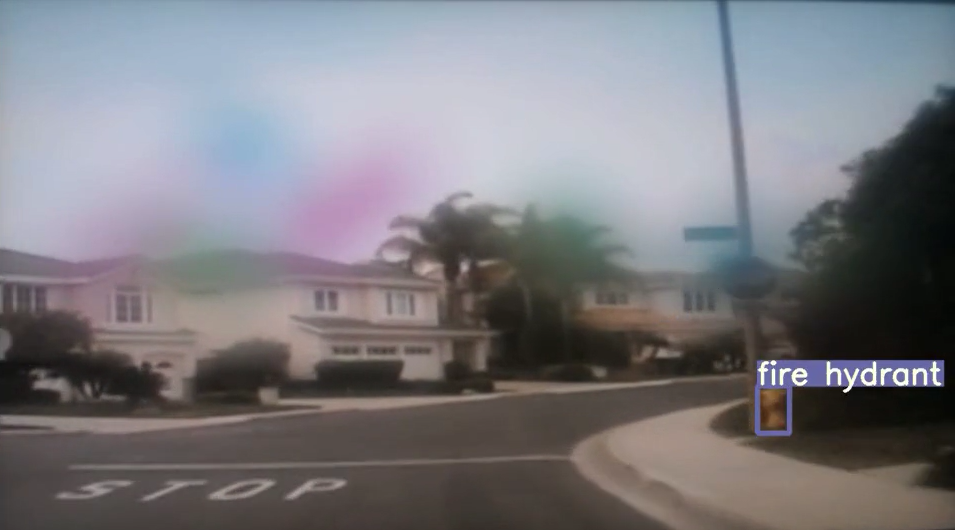}
    \end{subfigure}
    \begin{subfigure}{.19\linewidth}
        \centering
        \includegraphics[width=\linewidth]{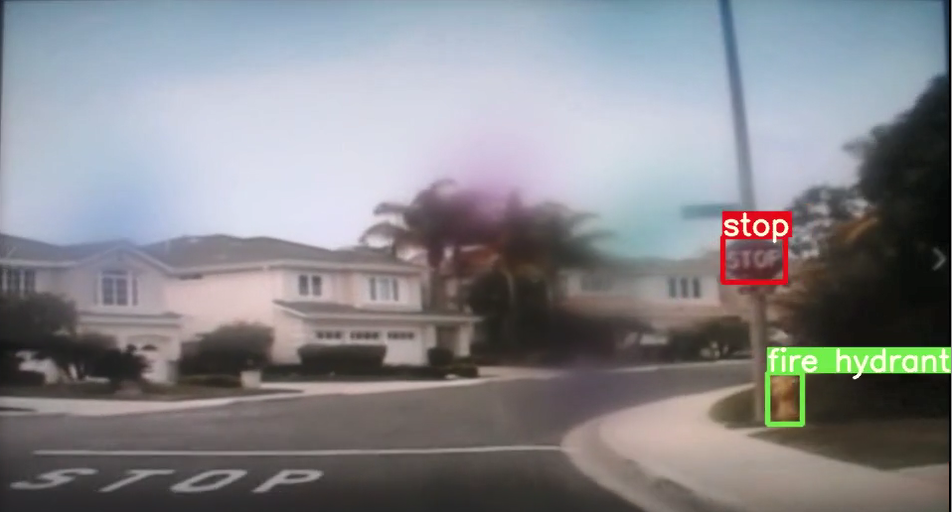}
    \end{subfigure}
    \begin{subfigure}{.19\linewidth}
        \centering
        \includegraphics[width=\linewidth]{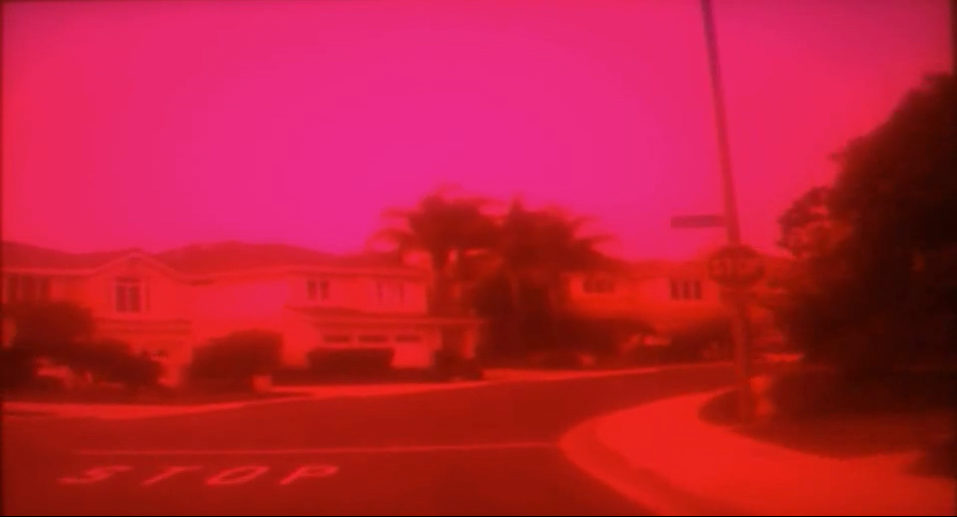}
    \end{subfigure}
    \begin{subfigure}{.19\linewidth}
        \centering
        \includegraphics[width=\linewidth]{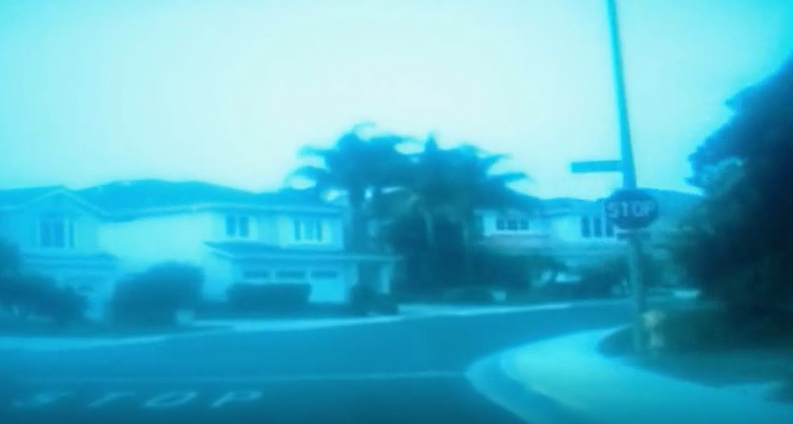}
    \end{subfigure}
    \begin{subfigure}{.19\linewidth}
        \centering
        \includegraphics[width=\linewidth]{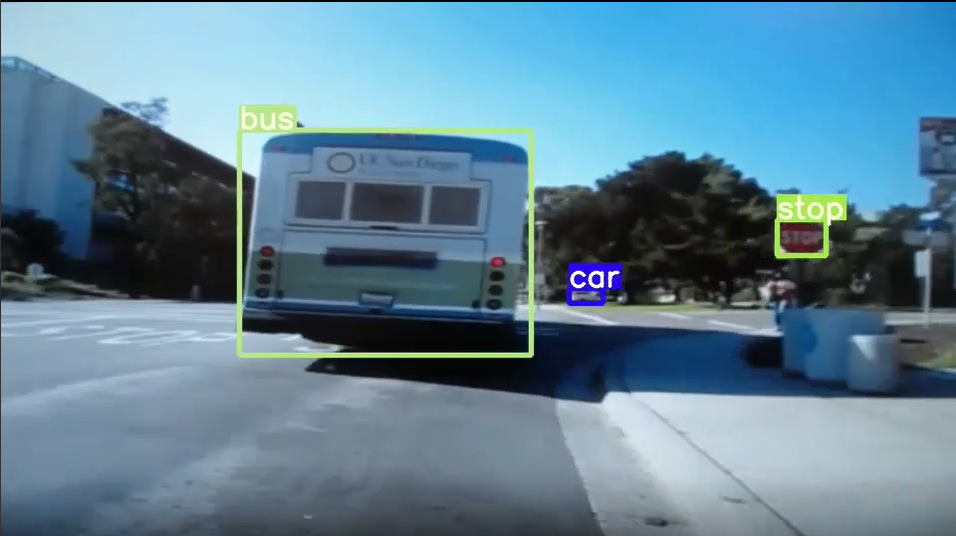}
    \end{subfigure}
    \begin{subfigure}{.19\linewidth}
        \centering
        \includegraphics[width=\linewidth]{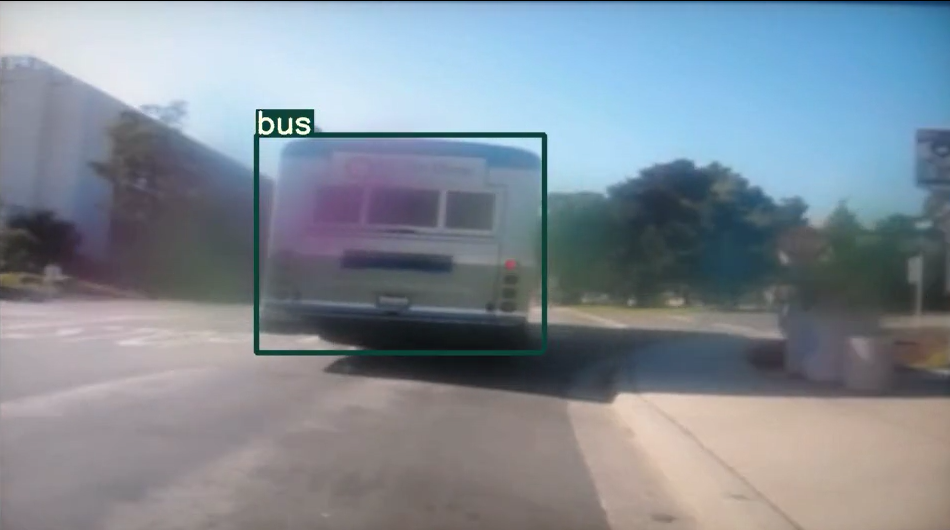}
    \end{subfigure}
    \begin{subfigure}{.19\linewidth}
        \centering
        \includegraphics[width=\linewidth]{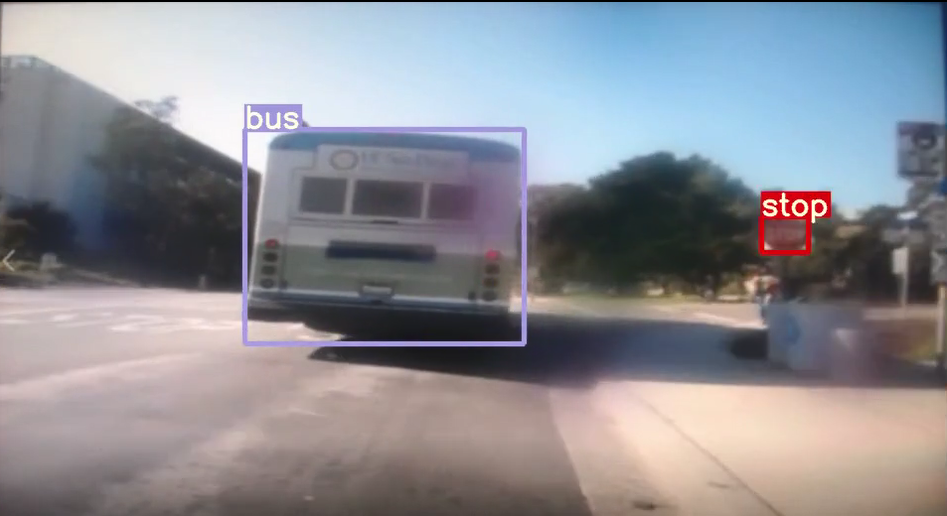}
    \end{subfigure}
    \begin{subfigure}{.19\linewidth}
        \centering
        \includegraphics[width=\linewidth]{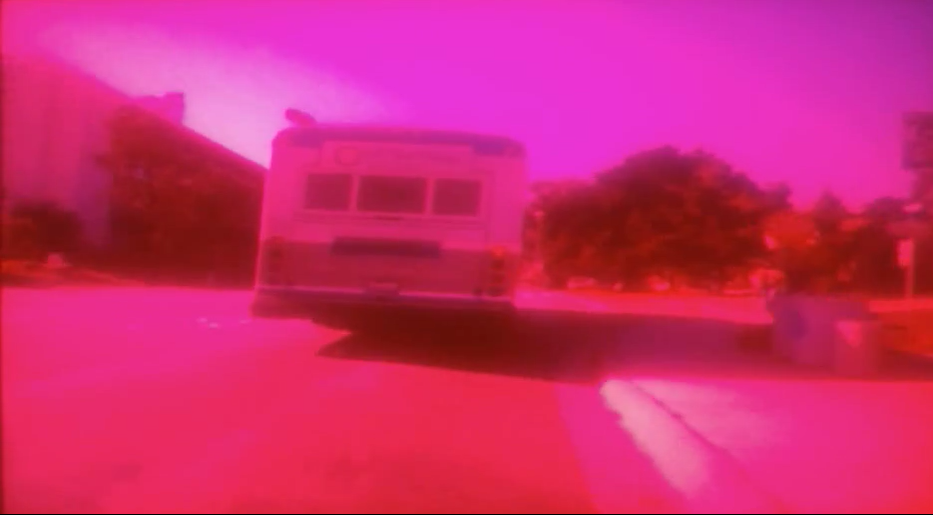}
    \end{subfigure}
    \begin{subfigure}{.19\linewidth}
        \centering
        \includegraphics[width=\linewidth]{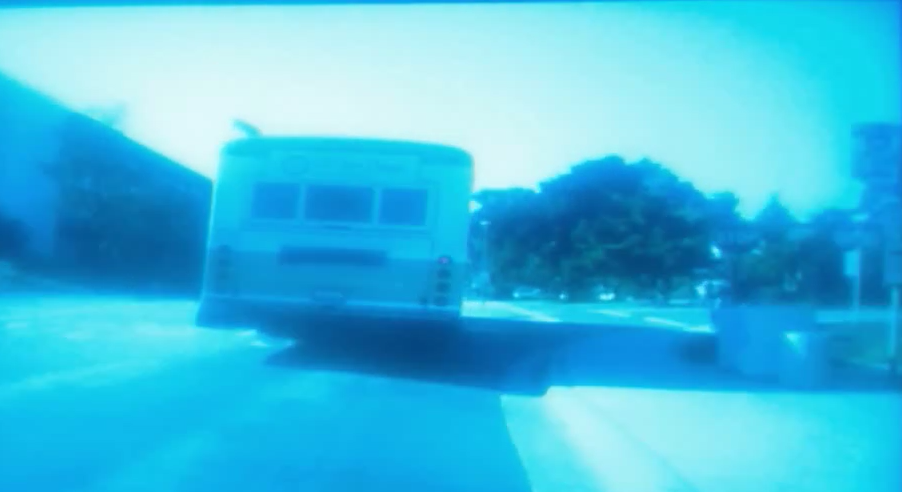}
    \end{subfigure}
    \caption{Examples of frames presented to the detection model and its detections when various patches are applied to the camera's lens. 
    Each row represents a single frame, and each column represents the use of a different patch.}
    \label{fig:attacks}
    \vspace{-0.4cm}
\end{figure*}

%% file: sections/conclusion.tex
\section{\label{sec:conclusion}Conclusion}
We presented a physically-realizable attack against state-of-the-art object detectors without the need to have direct access to the targeted object.
We crafted a translucent patch attached to the camera lens by the adversary with a unique design that considers real-world constraints and implemented a custom optimization process to achieve a successful attack in a real-world environment.
Our experiments demonstrated that it is possible to prevent a specific class from being detected while simultaneously allowing the detection of other classes.
Moreover, we showed that compared to the other evaluated attacks, our method has the best trade-off between the target class's misdetection and the untargeted classes' detection.

Our study highlighted the risk that autonomous cars' ADASs face from an adversary capable of simply applying a patch on their cameras.
Further research should focus on proposing countermeasures for such attacks, such as anomaly detection or active sensor checks.